\documentclass[journal]{IEEEtran}

\ifCLASSINFOpdf
\else
\fi

\usepackage{graphicx}
\usepackage{booktabs}
\usepackage{multirow}
\usepackage{amsmath}
\usepackage{amssymb}
\usepackage{balance}
\usepackage{cite}
\usepackage[colorlinks, citecolor = blue, urlcolor = magenta]{hyperref}

\begin{document}

\title{Image Dehazing as In-Context Tabular Inference}

\author{Le-Anh Tran \\
\thanks{tranlevision@gmail.com}
\thanks{\url{https://github.com/tranleanh/dehazetab}}
}

\maketitle

\begin{abstract}
Image dehazing is traditionally addressed through handcrafted physical priors or task-specific deep learning models, often requiring carefully designed architectures or extensive model training. Recent advances in tabular foundation models (TFMs), however, offer an alternative learning paradigm in which continuous relationships can be inferred directly from contextual examples without task-specific model training. Motivated by this capability, this paper investigates whether single-image dehazing can be effectively reformulated as an in-context tabular regression problem and introduces DehazeTab, a framework that estimates atmospheric transmission from structured image-patch representations. Specifically, overlapping patches extracted from hazy images are transformed into tabular samples using handcrafted haze-related descriptors, pretrained visual embeddings, or their hybrid combinations, while corresponding transmission patches are summarized as scalar regression targets. A pretrained TFM then performs in-context regression using labeled reference patches to predict transmission values for unseen images without any further fine-tuning. These predictions are spatially aggregated into a transmission map, refined through guided filtering, and incorporated into the atmospheric scattering model to recover haze-free images. Experiments on benchmark datasets demonstrate competitive restoration performance and reveal that physically motivated handcrafted features can outperform pretrained visual embeddings, while hybrid representations offer complementary benefits. Further analyses examine the effects of feature representations, patch selection, and contextual reference configurations on restoration quality. Beyond image dehazing, the findings highlight the potential of TFMs as an alternative paradigm for addressing low-level vision tasks through structured and in-context inference. 
\end{abstract}


%
\IEEEpeerreviewmaketitle

\section{Introduction}
\label{sec:introduction}

\IEEEPARstart{I}{mage} dehazing seeks to recover clear scene information obscured by atmospheric haze, which reduces contrast, alters colors, and suppresses structural details, compromising visual quality and downstream vision tasks. Yet single-image dehazing is inherently ill-posed: distinct clear scenes can produce similar hazy observations. Recovering a plausible scene therefore requires physical priors or learned constraints. The widely used atmospheric scattering model describes this degradation as:
\begin{equation}
\mathbf{I}(x) = \mathbf{J}(x)t(x) + \mathbf{A}\bigl(1-t(x)\bigr),
\label{eq:atmospheric_scattering_model}
\end{equation}
where $\mathbf{I}(x)$ and $\mathbf{J}(x)$ are the hazy observation and haze-free radiance, $\mathbf{A}$ denotes atmospheric light, and $t(x)$ is transmission. This physically interpretable formulation makes transmission estimation central to restoration. Conventional methods estimate transmission from natural-image statistics. Dark channel prior (DCP) \cite{he2010single}, for instance, exploits local intensity patterns, while saturation line prior (SLP) \cite{ling2023single} uses local saturation--brightness relationships to infer haze-related quantities. These approaches suggest an important observation: image patches contain structured cues from which transmission can be predicted as a continuous value. However, fixed priors may fail when scene characteristics violate their assumptions. Deep learning offers more flexible mappings, estimating transmission or directly restoring clear images from paired data \cite{cai2016dehazenet,li2017aod,qin2020ffa,tran2025encoder,song2023vision}. Despite impressive results, such models often depend on task-specific training and substantial paired data, and synthetic-to-real domain shifts remain challenging. Some weakly supervised, synthetic-to-real, zero-shot, and unpaired methods also seek to alleviate these limitations \cite{zhao2021refinednet,chen2021psd,dong2023semi,li2020zeroshot,li2021you,tran2026unpaired}.

\begin{table}[t]
\centering
\caption{Illustration of the patch-to-tabular formulation. Labeled patches constitute the context, while unseen patches are provided without transmission targets.}
\label{tab:patch_to_tabular}
\renewcommand{\arraystretch}{1.15}
\setlength{\tabcolsep}{4pt}
\begin{tabular}{c c c c c c c c}
\toprule
\multirow{2}{*}{\textbf{Type}} &
\multirow{2}{*}{\textbf{Patch}} &
\multicolumn{5}{c}{\textbf{Features}} &
\multirow{2}{*}{\textbf{Target}} \\
\cmidrule{3-7}
&
&
$\boldsymbol{f_1}$ &
$\boldsymbol{f_2}$ &
$\boldsymbol{f_3}$ &
$\boldsymbol{\cdots}$ &
$\boldsymbol{f_n}$ & \\
\midrule
\multirow{4}{*}{Context}
& $P_1$ & 0.61 & 0.18 & 0.43 & $\cdots$ & 0.52 & 0.78 \\
& $P_2$ & 0.73 & 0.11 & 0.57 & $\cdots$ & 0.66 & 0.61 \\
& $P_3$ & 0.45 & 0.27 & 0.31 & $\cdots$ & 0.48 & 0.84 \\
& $\vdots$ & $\vdots$ & $\vdots$ & $\vdots$ & $\ddots$ & $\vdots$ & $\vdots$ \\
\midrule
\multirow{3}{*}{Unseen}
& $P_{u,1}$ & 0.68 & 0.15 & 0.49 & $\cdots$ & 0.58 & \textcolor{red}{$\times$} \\
& $P_{u,2}$ & 0.79 & 0.09 & 0.63 & $\cdots$ & 0.71 & \textcolor{red}{$\times$} \\
& $\vdots$ & $\vdots$ & $\vdots$ & $\vdots$ & $\ddots$ & $\vdots$ & $\vdots$ \\
\bottomrule
\end{tabular}
\end{table}

Meanwhile, tabular foundation models (TFMs) \cite{gardner2024large,thomas2024retrieval,hollmann2025accurate} have introduced a different way to learn from structured data. Rather than optimizing a separate regressor for every task, a pretrained TFM can infer feature--target relationships in context, directly from labeled examples. This capability is particularly appealing for small- and medium-scale datasets, reducing the need for repeated training, hyperparameter tuning, and task-specific architecture design. It raises an interesting question: \emph{can image dehazing be solved through in-context tabular inference instead of training a dedicated network?}

Motivated by this question, the paper proposes \textbf{DehazeTab}, which reformulates transmission estimation as patch-level tabular regression. Each hazy-image patch becomes a row of structured features, paired with a scalar transmission target, as illustrated in Table~\ref{tab:patch_to_tabular}. Given labeled reference patches, a pretrained TFM predicts transmission for patches from an unseen image without task-specific fine-tuning. Overlapping predictions are aggregated into a dense transmission map, refined by guided filtering \cite{he2012guided} to preserve boundaries and reduce artifacts, and used in the atmospheric scattering model to recover the haze-free scene.
Accordingly, feature representation is crucial to this formulation. Handcrafted haze descriptors explicitly encode physical and local statistical cues; pretrained visual embeddings offer richer structural and semantic information; and hybrid representations may combine their strengths. DehazeTab therefore examines these complementary representations to understand what information supports effective in-context transmission prediction. In a nutshell, the main contributions include:
\begin{itemize}
    \item The \textbf{DehazeTab} framework is introduced to reformulate single-image dehazing as patch-level structured regression using a pretrained TFM and contextual examples.

    \item A patch-to-tabular pipeline is developed to transform image features into transmission predictions, followed by spatial aggregation and image reconstruction.

    \item Handcrafted, pretrained visual, and hybrid feature representations are systematically investigated to examine their contributions to the framework's performance.

    \item Benchmark comparisons and ablation studies are conducted to assess the feasibility and limitations of the proposed framework, highlighting its potential for broader image restoration applications.
\end{itemize}

The goal is not to claim state-of-the-art performance, but to examine whether structured in-context inference offers a viable alternative perspective on image restoration. Section~\ref{sec:relatedwork} reviews related work; Section~\ref{sec:method} details the DehazeTab framework; Section~\ref{sec:experiments} presents experiments and analysis; and Section~\ref{sec:conclusions} concludes the paper.

\begin{figure*}
    \centering
    \includegraphics[width=0.99\textwidth]{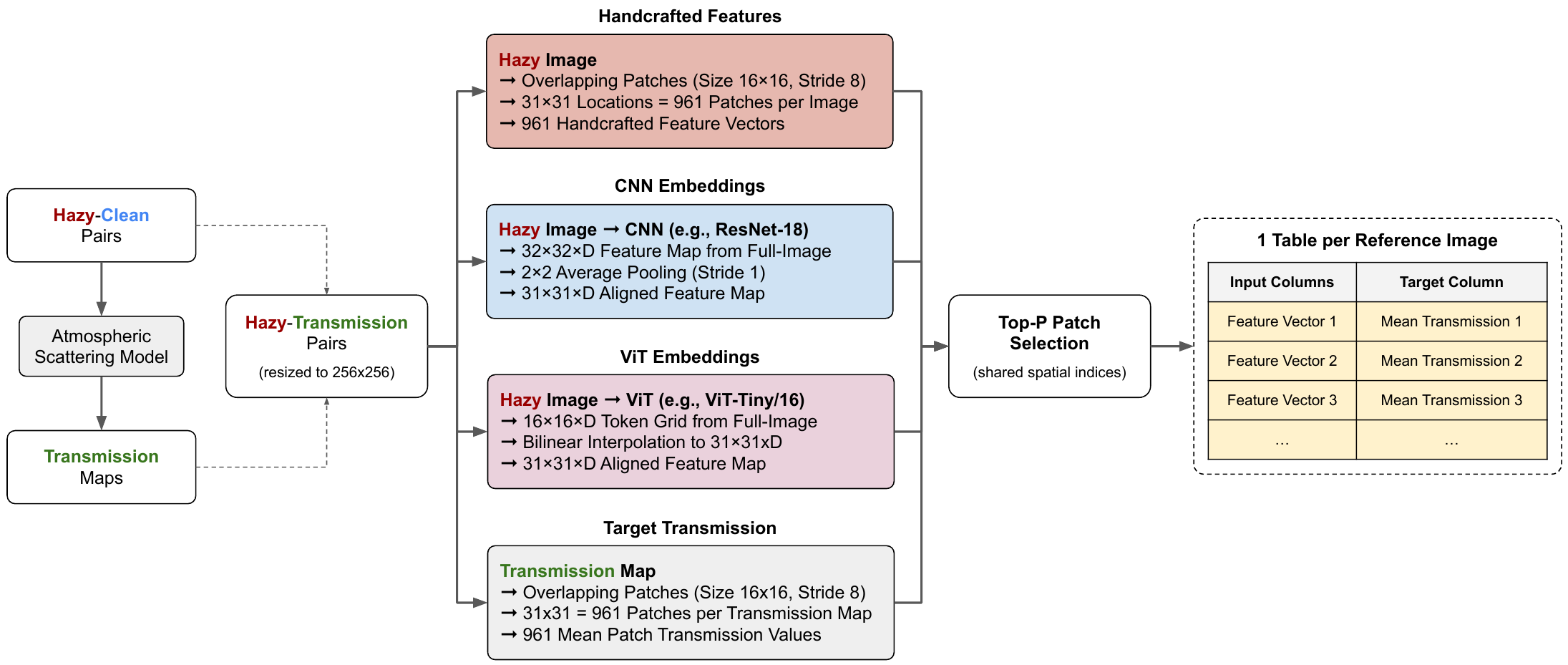}
    \caption{Offline preparation of spatially aligned patch features and transmission targets for contextual table construction.}
    \label{fig:table_preparation}
\end{figure*}

\begin{table*}[t]
\centering
\caption{Regression input configurations for transmission estimation using tabular foundation models.}
\label{tab:research_questions}
\begin{tabular}{l p{5.8cm} p{5.8cm}}
\toprule
\textbf{Type} & \textbf{Description } & \textbf{Research Question} \\
\midrule
(i) Handcrafted features
& DCP, statistical, and texture features
& Are physically motivated features effective for this reformulated task? \\
(ii) Local embedded features
& Features from a pretrained CNN (e.g., ResNet)
& Do local and hierarchical visual representations provide better transmission cues? \\
(iii) Global embedded features
& Features from a pretrained ViT (e.g., ViT-Tiny)
& Does global contextual information improve transmission estimation? \\
(iv) Hybrid features
& Handcrafted + embedded features
& Do handcrafted and learned features provide complementary information? \\
\bottomrule
\end{tabular}
\end{table*}

\begin{figure*}
    \centering
    \includegraphics[width=0.99\textwidth]{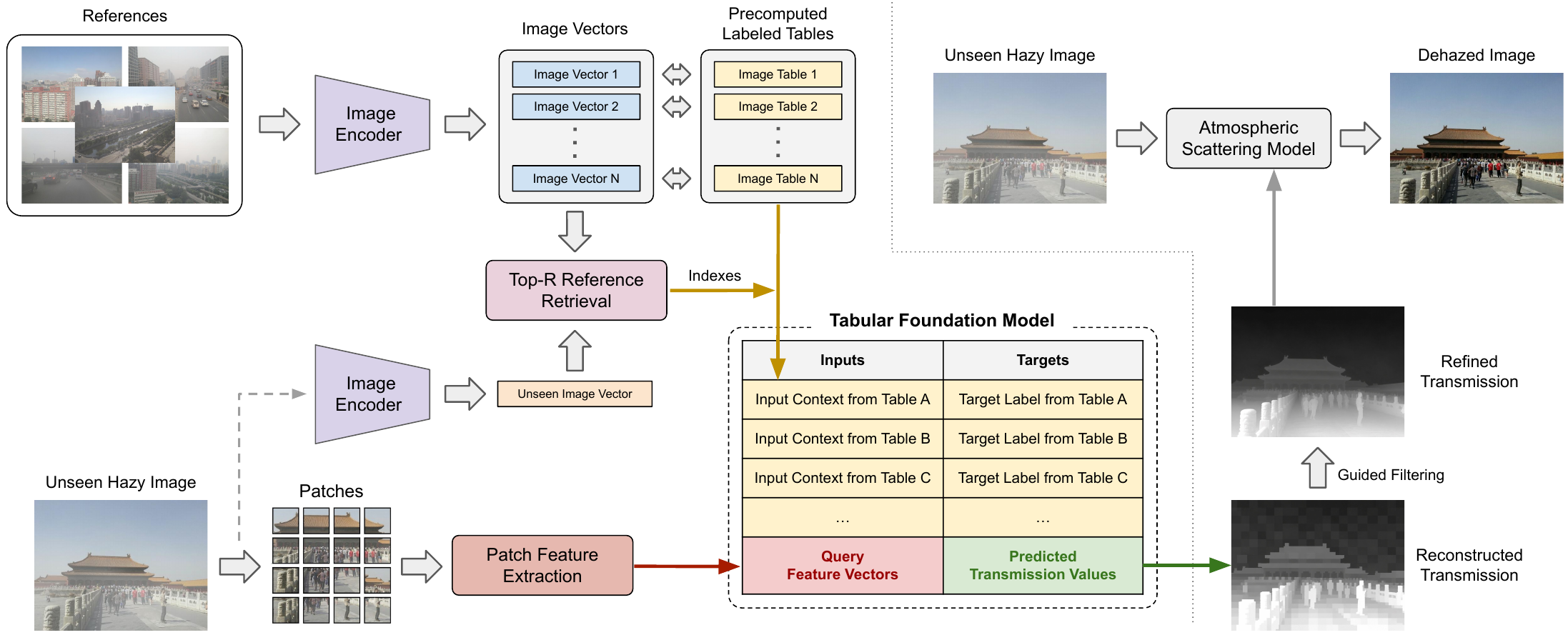}
    \caption{Reference retrieval and in-context inference process of DehazeTab.}
    \label{fig:framework}
\end{figure*}

\section{Related Work}
\label{sec:relatedwork}

\subsection{Prior-Based Approaches}

Single-image dehazing is commonly formulated using the atmospheric scattering model, which relates a hazy observation to scene radiance, transmission, and atmospheric light. Since recovering these quantities from one image is ill-posed, traditional methods rely on statistical priors. The influential DCP \cite{he2010single} estimates transmission from the observation that most haze-free outdoor patches contain low-intensity pixels in at least one color channel. 
Subsequent approaches exploit other local color relationships. SLP \cite{ling2023single} models local saturation-brightness relationships to estimate transmission. 
RSVT \cite{tran2024single} uses saturation-value relationships and relatively stable hue information to improve restoration in bright and sky regions. Together, these studies establish an important insight: local image regions contain structured, physically meaningful cues for inferring haze-related quantities.
While physically interpretable and free of data-driven training, such methods can fail when their assumptions do not hold, and their inference rules remain largely prescribed by handcrafted assumptions rather than adapted from contextual examples.

\subsection{Learning-Based Approaches}

Deep learning enables more flexible dehazing by learning transmission estimates or direct hazy-to-clear mappings from data. DehazeNet \cite{cai2016dehazenet} predicts transmission with a convolutional network, whereas AOD-Net \cite{li2017aod} supports end-to-end image reconstruction through a reformulation of the scattering model. Subsequent architectures, including EDN-GTM \cite{tran2022novel}, FFA-Net \cite{qin2020ffa}, DPTE-Net \cite{tran2025distilled}, and DehazeFormer \cite{song2023vision}, improve restoration through guided transmission estimation, feature fusion, distillation, and attention-based representations. Despite strong benchmark performance, supervised approaches depend on paired training data, which are difficult to collect under real atmospheric conditions. Models trained predominantly on synthetic haze also face potential domain shifts.

To reduce these dependencies, RefineDNet \cite{zhao2021refinednet} employs weak supervision, PSD \cite{chen2021psd} adapts synthetic-pretrained models using physical priors, and SDA-GAN \cite{dong2023semi} addresses synthetic-to-real discrepancies through semi-supervised domain alignment. Other methods avoid paired training collections: ZID \cite{li2020zeroshot} and YOLY \cite{li2021you} estimate atmospheric components through zero-shot or image-specific optimization, while UID-KAT \cite{tran2026unpaired} investigates unpaired restoration through latent feature transformation and adversarial learning. Although these directions relax supervision requirements, they generally retain specialized neural architectures, task-specific learning procedures, or image-specific optimization. This motivates exploring whether dehazing can instead be performed using a pretrained, general-purpose predictor conditioned on representative examples.

\subsection{Tabular Foundation Models for Vision Problems}

Tabular foundation models (TFMs) offer a distinct learning paradigm: rather than fitting a new predictor for every dataset, a pretrained model can infer feature-target relationships from labeled examples supplied as context. TabPFN \cite{hollmann2025accurate}, pretrained on synthetic tabular tasks, performs classification and regression through in-context inference without task-specific gradient-based optimization. Related advances include TABULA-8B \cite{gardner2024large}, which investigates zero- and few-shot transfer across heterogeneous tables through language modeling, and LoCalPFN \cite{thomas2024retrieval}, which extends tabular prediction through retrieval and fine-tuning. These developments highlight the potential of TFMs to generalize statistical inference capabilities across structured data and in-context prediction.

Nevertheless, TFM research has largely concentrated on conventional tabular datasets, while their applicability to dense, low-level vision tasks remains comparatively unexplored. Addressing this gap, DehazeTab represents one of the early investigations of TFMs for single-image dehazing. 
It transforms overlapping image patches into structured feature vectors, uses contextual transmission labels for patch-level regression, and reconstructs haze-free images through the atmospheric scattering model. Unlike conventional dehazing networks, the proposed framework places the learning task in the tabular feature-to-transmission relationship, allowing a pretrained TFM to infer haze characteristics without dehazing-specific weight updates. Beyond restoration accuracy, this formulation explores a broader direction for connecting general-purpose in-context learning with image restoration.

\section{Methodology}
\label{sec:method}

\subsection{Overview and Problem Formulation}
\label{subsec}

The core idea is to reformulate atmospheric transmission estimation as an in-context tabular regression problem, offering an alternative to conventional task-specific dehazing networks. Given a hazy image $\mathbf{I}$ and its corresponding transmission map $\mathbf{T}$ from a reference image pair, overlapping image patches are represented by compact feature vectors $\mathbf{x}_i\in\mathbb{R}^{d}$, while their aligned transmission patches provide scalar regression targets $y_i$. The resulting contextual dataset is defined as:
\begin{equation}
\mathcal{D} = \{(\mathbf{x}_i, y_i)\}_{i=1}^{N},
\end{equation}
where each row represents an image patch and its associated transmission value, as illustrated in Table~\ref{tab:patch_to_tabular}. Unlike conventional regressors that require task-specific optimization, a pretrained TFM directly conditions on these labeled examples to predict transmission values for unseen image patches without updating model parameters.

A key consideration in this formulation is how image patches should be represented for effective transmission inference. Accordingly, DehazeTab investigates four feature configurations, summarized in Table~\ref{tab:research_questions}: (i) handcrafted haze-related descriptors capturing physical and statistical cues, (ii) CNN-based embeddings encoding local and hierarchical visual structures, (iii) ViT-based embeddings incorporating broader contextual information \cite{dosovitskiy2021image}, and (iv) hybrid representations combining handcrafted and learned features. These configurations systematically examine whether physically motivated features, learned visual representations, or their combination provide more effective transmission cues. Finally, the predicted patch-level transmission values are spatially aggregated and refined to construct a dense transmission map for haze-free image reconstruction.

\subsection{Contextual Sample Construction}
\label{subsec:training_sample_prepration}

\textbf{Transmission targets.} Contextual examples are derived from paired hazy and haze-free reference images. Since atmospheric light $\mathbf{A}$ is not provided, it is estimated using DCP \cite{he2010single}: candidate pixels are identified from high dark-channel responses, and the brightest corresponding hazy-image pixel supplies the atmospheric-light estimate $\hat{\mathbf{A}}$. Given an aligned hazy--clear pair and the atmospheric scattering model in Eq.~(\ref{eq:atmospheric_scattering_model}), a channel-shared transmission estimate is:
\begin{equation}
    \hat{t}(x)=
    \frac{
    \bigl(\mathbf{J}(x)-\hat{\mathbf{A}}\bigr)^T
    \bigl(\mathbf{I}(x)-\hat{\mathbf{A}}\bigr)
    }{
    \left\|\mathbf{J}(x)-\hat{\mathbf{A}}\right\|_2^2+\epsilon
    },
\end{equation}
where $\epsilon$ ensures numerical stability, and the estimate is then clipped to $[0,1]$.

\textbf{Patch-level targets.}
For each reference image, overlapping patches are extracted from the hazy image and its estimated transmission map using the same spatial coordinates. For patch region $\Omega_i$, the scalar regression target is defined as its mean transmission:
\begin{equation}
\bar{t}_i=\frac{1}{|\Omega_i|}\sum_{x\in\Omega_i}\hat{t}(x).
\end{equation}
The corresponding hazy patch is represented by a feature vector $\mathbf{x}_i$, forming a labeled tabular sample $(\mathbf{x}_i,\bar{t}_i)$. By using consistent patch coordinates for feature extraction and transmission-target construction, each feature vector is associated with the mean transmission of its corresponding spatial region, ensuring spatial alignment between the inputs and regression targets.

\textbf{Context patch selection.}
Retaining all extracted patches would introduce substantial redundancy and increase the computational cost of in-context inference. To construct a compact and representative context table, only the top $P$ patches are selected from each reference image. The primary strategy employs an informativeness score based on local contrast, intensity range, gradient strength, entropy, dark-channel variation, and saturation variation, together with a spatial-distance constraint to reduce near-duplicates. Alternatively, clustering-based selection identifies representative patches in feature space, promoting sample diversity beyond predefined local statistics. The effects of those selection strategies on dehazing performance are examined in Section~\ref{subsec:additional_analysis}.

\subsection{Patch Feature Representations}

\subsubsection{Handcrafted Haze-Related Features}
\label{subsec:handcrafted_features}

The first representation tests whether physically meaningful local statistics provide sufficient information for transmission regression. Each hazy patch $\Omega_i$ is characterized by RGB and grayscale intensity statistics, saturation and brightness, inter-channel differences, contrast, gradients, edge and texture descriptors, entropy, and Laplacian variance. Dark-channel statistics are additionally included because of their established relationship to haze. The resulting handcrafted feature vector is expressed as:
\begin{equation}
\mathbf{x}_i^{H}=
[\mathbf{f}_{\mathrm{color}},
\mathbf{f}_{\mathrm{intensity}},
\mathbf{f}_{\mathrm{DCP}},
\mathbf{f}_{\mathrm{contrast}},
\mathbf{f}_{\mathrm{texture}}],
\label{eq:handcrafted_features}
\end{equation}
which defines the \textbf{DehazeTab$_H$} variant.

\subsubsection{Pretrained Visual Embeddings}
\label{subsec:pretrained_embeddings}

The second representation investigates whether features learned for general visual understanding can provide useful cues for atmospheric transmission estimation. Rather than independently encoding each small image patch, a frozen pretrained encoder processes the entire hazy image to preserve broader visual context. Given a $256 \times 256$ input RGB image:
\begin{equation}
    \mathbf{I}\in\mathbb{R}^{256\times256\times3},
\end{equation}
the encoder produces an intermediate feature map, which is subsequently aligned with the overlapping patch grid used for transmission estimation.

For CNN-based features, a pretrained ResNet-18 encoder truncated after \textit{layer 2} is employed, producing:
\begin{equation}
    \mathbf{F}_{CNN}=E_{CNN}(\mathbf{I}),\qquad
    \mathbf{F}_{CNN}\in\mathbb{R}^{32\times32\times128}.
\end{equation}
Then a $2\times2$ average-pooling operation with stride 1 is applied:
\begin{equation}
\begin{aligned}
    \widetilde{\mathbf{F}}_{CNN}
    &= \operatorname{AvgPool}_{2\times2,s=1}
    (\mathbf{F}_{CNN}), \\
    \widetilde{\mathbf{F}}_{CNN}
    &\in \mathbb{R}^{31\times31\times128}.
\end{aligned}
\end{equation}
For ViT-based features, a pretrained ViT-Tiny/16 is adopted to incorporate broader contextual information through self-attention. Unlike the CNN encoder, ViT-Tiny/16 employs non-overlapping $16\times16$ patch embeddings with a stride of 16. Consequently, a $256\times256$ input naturally produces a $16\times16$ spatial token grid, excluding the class token:
\begin{equation}
    \mathbf{F}_{ViT}=E_{ViT}(\mathbf{I}),\qquad
    \mathbf{F}_{ViT}\in\mathbb{R}^{16\times16\times192}.
\end{equation}
This resolution follows directly from the patch-embedding configuration of the pretrained model, rather than from an arbitrary choice of spatial downsampling. Obtaining a denser native token grid would require modifying the input resolution or adopting a different patch-embedding configuration, potentially increasing computational cost or changing the pretrained representation. Therefore, bilinear interpolation with aligned corners is employed to map the contextualized spatial tokens directly onto the overlapping patch grid:
\begin{equation}
\begin{aligned}
    \widetilde{\mathbf{F}}_{ViT}
    &= \operatorname{Interp}_{31\times31}
    (\mathbf{F}_{ViT}), \\
    \widetilde{\mathbf{F}}_{ViT}
    &\in \mathbb{R}^{31\times31\times192}.
\end{aligned}
\end{equation}

Both feature representations are thus spatially aligned with the $31\times31$ grid of overlapping $16\times16$ image patches extracted with stride 8. For either encoder, the embedding associated with patch $\Omega_i$ is obtained as:
\begin{equation}
    \mathbf{x}_i^{E}
    =\widetilde{\mathbf{F}}_{u_i,v_i},
    \qquad
    \mathbf{x}_i^{E}\in\mathbb{R}^{D},
    \label{eq:learned_visual_embeddings}
\end{equation}
where $(u_i,v_i)$ denotes the spatial grid coordinates of patch $\Omega_i$, and $D$ is the corresponding feature dimension. The same extraction procedure is applied during reference preparation and query inference. 
The CNN representation provides hierarchical spatial features, whereas the ViT representation captures broader contextual relationships through self-attention. These alternatives define \textbf{DehazeTab$_{CNN}$} and \textbf{DehazeTab$_{ViT}$}, respectively, allowing their effectiveness for patch-level transmission estimation to be systematically evaluated.

\subsubsection{Hybrid Feature Representation}
\label{subsec:hybrid_representations}

Handcrafted descriptors explicitly capture haze-related statistics, while pretrained embeddings encode richer visual structures. To investigate their complementarity, the two representations are concatenated:
\begin{equation}
    \mathbf{x}_i^{Hyb} = \left[\mathbf{x}_i^{H},\mathbf{x}_i^{E}\right].
\end{equation}
CNN-based embeddings are adopted for this hybrid representation because their local inductive biases are well suited to capturing haze-related spatial cues, whereas ViT-based embeddings emphasize broader contextual relationships, which may be less informative for local transmission estimation. This choice is further supported by the quantitative comparisons in Section~\ref{subsec:quantitative_results}. The hybrid construction uses the same selected patch coordinates for both feature types, ensuring spatial correspondence between handcrafted descriptors and learned embeddings. This configuration is denoted \textbf{DehazeTab$_{Hybrid}$}. The transmission targets and TFM regression mechanism remain unchanged across variants, allowing the effect of feature representation to be evaluated independently. The offline feature preparation process is illustrated in Figure~\ref{fig:table_preparation}.

\subsection{In-Context Transmission Inference and Reconstruction}
\label{subsec:in_context_inference}

\textbf{Reference retrieval.} For each unseen hazy image, $R$ reference images are identified using image-level embeddings from a pretrained ResNet encoder. The most similar references supply their previously constructed labeled patch tables, which are combined into $\mathcal{D}_{\mathrm{context}}$, providing scene-relevant context for TFM with a contextual table containing $P \times R$ labeled samples.
The impacts of different numbers of references $R$ and selected patches $P$ are examined in Section \ref{subsec:additional_analysis}.

\textbf{In-context regression.} The unseen image is represented using the same feature extraction scheme as the references. Given an unlabeled patch vector $\mathbf{x}_j$, the pretrained TFM predicts its transmission by conditioning on the labeled context:
\begin{equation}
    \hat{y}_j=f_{\mathrm{TFM}}\!\left(\mathbf{x}_j\mid\mathcal{D}_{\mathrm{context}}\right).
\end{equation}
Thus, the relationship between patch features and transmission is inferred from contextual examples, rather than learned by training a dehazing-specific regressor. TabPFN \cite{hollmann2025accurate} is adopted as the TFM in this study due to its strong regression performance on small- and medium-scale tabular datasets and its proven ability to perform in-context inference.

\textbf{Haze-free image recovery.} Predicted patch-level transmission values are projected back onto their spatial regions and aggregated across overlapping patches to form a dense, coarse transmission map. Guided filtering \cite{he2012guided} then refines this map to reduce patch artifacts while preserving structural boundaries. The refined transmission is resized to the original image resolution and combined with the estimated atmospheric light to recover the haze-free scene. The full process of reference retrieval and in-context inference of DehazeTab is illustrated in Figure~\ref{fig:framework}.

\begin{table*}[ht]
\centering
\caption{Performance comparisons on the SOTS-Outdoor and HSTS datasets (\textbf{bold}: best, \underline{underline}: second-best).}
\begin{tabular}{llccccc}
\toprule
\multirow{2}{*}{\text{Method}} & \multirow{2}{*}{\text{Publication}} & \multirow{2}{*}{\text{Type}} & \multicolumn{2}{c}{\text{SOTS-Outdoor}} & \multicolumn{2}{c}{\text{HSTS}} \\
\cmidrule(r){4-5} \cmidrule(r){6-7}
& & & \text{PSNR$\uparrow$} & \text{SSIM$\uparrow$} & \text{PSNR$\uparrow$} & \text{SSIM$\uparrow$} \\
\midrule
DCP \cite{he2010single} & TPAMI'11 & Prior & 18.38 & 0.819 & 17.01 & 0.803 \\
BCCR \cite{meng2013efficient} & ICCV'13 & Prior & 15.71 & 0.769 & 15.21 & 0.747 \\
JEEF \cite{kaplan2023real} & JVCIR'23 & Prior & 17.51 & 0.857 & 17.61 & 0.861 \\
RSVT \cite{tran2024single} & PCS'24 & Prior & 22.31 & 0.903 & 21.53 & 0.887 \\
\midrule
ZID \cite{li2020zeroshot} & TIP'20 & Zero-shot & 20.27 & 0.878 & 22.65 & 0.901 \\
YOLY \cite{li2021you} & IJCV'21 & Unsupervised & 20.39 & 0.889 & 21.02 & 0.905 \\
PSD \cite{chen2021psd} & CVPR'21 & Unsupervised & 20.49 & 0.844 & 19.37 & 0.824 \\
RefineDNet \cite{zhao2021refinednet} & TIP'21 & Weakly-supervised & 20.81 & 0.877 & 20.66 & 0.890 \\
USID-Net \cite{li2022usid} & TMM'22 & Unsupervised & 23.89 & 0.919 & 25.76 & \underline{0.923} \\
HTFA \cite{li2022haze} & KBS'22 & Unsupervised & 17.45 & 0.704 & 17.98 & 0.695 \\
VQDUID \cite{yang2023visual} & NN'23 & Unsupervised & 22.53 & 0.875 & 22.25 & 0.847 \\
SGDRL \cite{jia2024self} & NN'24 & Unsupervised & 23.28 & 0.920 & 22.01 & 0.888 \\
HDUD \cite{li2024hdud} & NCAA'24 & Unsupervised & 24.18 & 0.869 & 25.88 & 0.922 \\
UCL-Dehaze \cite{wang2024ucl} & TIP'24 & Unsupervised & 24.23 & 0.914 & \underline{26.03} & 0.919 \\
UID-KAT \cite{tran2026unpaired} & PR'26 & Unsupervised & 24.57 & 0.917 & \textbf{27.16} & \textbf{0.930} \\
\midrule
DehazeTab$_{H}$ & Proposed & In-context & \underline{24.71} & \underline{0.940} & 23.45 & 0.918 \\
DehazeTab$_{CNN}$ & Proposed & In-context & 23.74 & 0.898 & 23.27 & 0.893 \\
DehazeTab$_{ViT}$ & Proposed & In-context & 23.06 & 0.884 & 21.18 & 0.858 \\
DehazeTab$_{Hybrid}$ & Proposed & In-context & \textbf{24.87} & \textbf{0.942} & 23.36 & 0.916 \\
\bottomrule
\end{tabular}
\label{tab:quantitative_results}
\end{table*}

\section{Experiments and Analyses}
\label{sec:experiments}

\subsection{Experimental Setup}
\label{subsec:experimental_setup}

\textbf{Datasets and evaluation protocol.} Experiments are conducted with two subsets of RESIDE database \cite{li2018benchmarking}: SOTS-Outdoor (500 hazy/clean pairs) and HSTS (10 pairs). For contextual reference construction, the 500 SOTS-Outdoor pairs are processed as described in Section~\ref{subsec:training_sample_prepration}. When evaluating a SOTS-Outdoor image, that image is excluded from its candidate reference pool. This is a leave-one-image-out contextual evaluation on SOTS-Outdoor, not a fully independent training/test split. The HSTS experiments provide a separate evaluation subset, using the full available SOTS-Outdoor reference pool.

\textbf{Context construction and retrieval.}
Reference images are processed at $256\times256$ resolution to estimate transmission and construct patch-level contextual samples. 
Overlapping $16\times16$ patches with a stride of 8 yield 961 candidate patches per image, from which 100 representative patches ($P=100$) are selected by default using the informativeness-based strategy described in Section~\ref{subsec:training_sample_prepration} to construct compact contextual tables. During query inference, all 961 overlapping patches are retained for transmission prediction, enabling spatial aggregation of neighboring estimates to reduce block artifacts and improve the continuity of the reconstructed transmission map.
A pretrained ResNet-50 encoder extracts image-level embeddings for reference retrieval. For each query image, the five most similar eligible references ($R=5$) are retrieved, and their precomputed patch tables are combined to form a contextual dataset containing $P\times R=500$ samples. This retrieval strategy avoids redundant feature extraction for reference images. The resulting transmission map is subsequently resized to the original image resolution for restoration.

\begin{figure*}[t]
    \centering
    \includegraphics[width=0.98\textwidth]{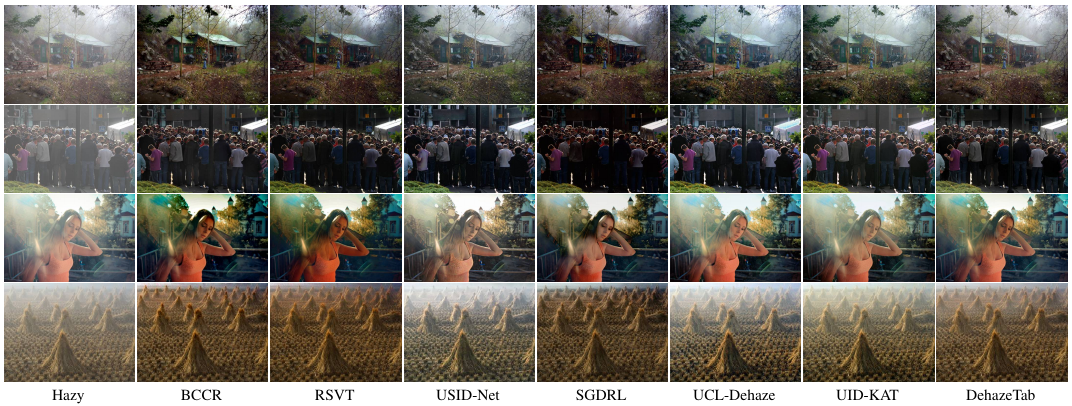}
    \caption{Typical visual results on natural hazy scenes.}
    \label{fig:results_natural} 
\end{figure*}

\textbf{Implementation and metrics.}
All variants employ the pretrained TabPFN-3.5 \cite{hollmann2025accurate}, with identical inference settings and normalized tabular features. Four feature representations are evaluated: DehazeTab$_H$, DehazeTab$_{CNN}$, DehazeTab$_{ViT}$, and DehazeTab$_{Hybrid}$. Reconstruction quality on paired benchmarks is assessed using PSNR and SSIM, where higher values indicate better performance. For real-world evaluation, NIQE and BRISQUE are employed to assess the perceptual quality of restored images without requiring haze-free ground truths; this evaluation is conducted on four representative natural hazy images commonly used in existing dehazing studies (shown in Figure \ref{fig:results_natural}), with lower NIQE and BRISQUE scores indicating better perceptual quality. Experiments are conducted on a Linux environment with an Intel Xeon Silver 4216 CPU 2.10 GHz and an NVIDIA GeForce RTX 3090 GPU 24 GB.

\subsection{Quantitative Comparisons}
\label{subsec:quantitative_results}

Table~\ref{tab:quantitative_results} compares DehazeTab against representative prior-based, zero-shot, unsupervised, and weakly supervised dehazing methods on SOTS-Outdoor and HSTS. The proposed variants are categorized as \emph{in-context}, since they use labeled reference examples at inference time, but do not fine-tune TabPFN for dehazing.

\begin{figure*}[t]
    \centering
    \includegraphics[width=0.98\textwidth]{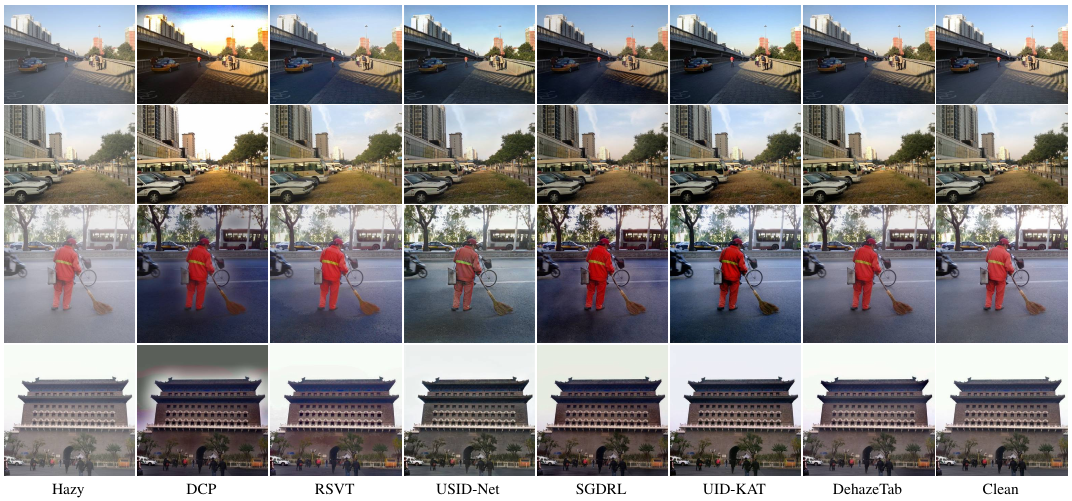}
    \caption{Typical visual results on SOTS-Outdoor.}
    \label{fig:results_sots} 
\end{figure*}

\begin{figure*}[t]
    \centering
    \includegraphics[width=0.98\textwidth]{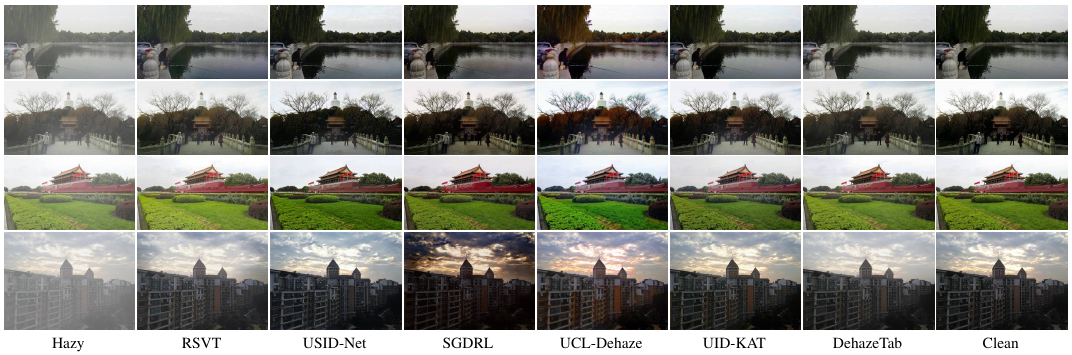}
    \caption{Typical visual results on HSTS.}
    \label{fig:results_hsts} 
\end{figure*}

\textbf{Effect of feature representation.}
On SOTS-Outdoor, the handcrafted variant achieves 24.71 dB PSNR and 0.940 SSIM, outperforming the CNN (23.74/0.898) and ViT (23.06/0.884) variants. Notably, CNN-based embeddings consistently outperform ViT-based embeddings across both benchmarks. This may be attributed to the stronger local inductive biases of CNNs, which help preserve haze-sensitive cues. In contrast, the broader contextual representations learned by ViTs may be less effective for patch-level transmission prediction in the proposed framework. These findings provide empirical support for adopting CNN-based embeddings in the hybrid representation. 
By combining handcrafted descriptors with CNN features, the hybrid variant achieves the strongest SOTS-Outdoor performance among the four configurations, reaching 24.87 dB PSNR and 0.942 SSIM, although the improvement over handcrafted features is modest. On HSTS, the handcrafted variant achieves 23.45 dB and 0.918 SSIM, slightly outperforming the hybrid variant (23.36/0.916), while the CNN and ViT variants obtain 23.27/0.893 and 21.18/0.858, respectively. Overall, these results highlight the effectiveness of physically motivated haze-related descriptors and suggest that incorporating generic pretrained visual embeddings does not necessarily improve transmission estimation, although combining handcrafted and learned features can provide complementary benefits.
The performance difference may also be influenced by the native spatial resolutions of the extracted representations, as CNN features originate from a denser spatial grid, whereas ViT features are interpolated from a coarser token grid. Thus, the observed advantage of CNN embeddings may reflect both their local inductive biases and their finer spatial sampling.

\begin{table}[t]
\centering
\caption{Performance comparisons on natural scenes (\textbf{bold}: best, \underline{underline}: second-best).}
\begin{tabular}{lccc}
\toprule
\multirow{2}{*}{\text{Method}} & \multirow{2}{*}{\text{Type}} & \multicolumn{2}{c}{\text{Natural Images}} \\
\cmidrule(r){3-4}
& & \text{NIQE$\downarrow$} & \text{BRISQUE$\downarrow$} \\
\midrule
DCP \cite{he2010single} & Prior & 3.21 & 15.64 \\
BCCR \cite{meng2013efficient} & Prior & 3.85 & 21.74 \\
RSVT \cite{tran2024single} & Prior & \textbf{3.08} & 14.66 \\
USID-Net \cite{li2022usid} & Unsupervised & 3.32 & 24.95 \\
SGDRL \cite{jia2024self} & Unsupervised & 3.21 & 15.41 \\
UCL-Dehaze \cite{wang2024ucl} & Unsupervised & 3.19 & \textbf{11.24} \\
UID-KAT \cite{tran2026unpaired} & Unsupervised & 3.29 & \underline{13.90} \\
\midrule
DehazeTab$_{H}$ & In-context & \underline{3.10} & 14.56 \\
DehazeTab$_{CNN}$ & In-context & \textbf{3.08} & 15.32 \\
DehazeTab$_{ViT}$ & In-context & 3.11 & 16.10 \\
DehazeTab$_{Hybrid}$ & In-context & \textbf{3.08} & 14.59 \\
\bottomrule
\end{tabular}
\label{tab:results_natural}
\end{table}

\textbf{Comparison with established methods.} 
DehazeTab produces competitive results despite relying on a general-purpose tabular regressor, its best SOTS-Outdoor result exceeds the reported PSNR of the compared methods such as UID-KAT or UCL-Dehaze, whereas on HSTS, they achieve higher PSNR than any DehazeTab variant. These comparisons support the feasibility of the proposed formulation, but do not establish a consistent performance advantage across datasets. 
In addition, because the compared methods differ in their supervision requirements, these results should be interpreted as indicative performance comparisons rather than evaluations under strictly identical data-access conditions.
The primary objective is to examine the potential of in-context tabular inference for image restoration, rather than to claim state-of-the-art accuracy.

\textbf{Natural-image evaluation.} 
Table~\ref{tab:results_natural} reports the NIQE and BRISQUE scores on a set of natural hazy scenes, where lower values indicate better perceptual quality. DehazeTab achieves competitive performance compared with established prior-based and unsupervised methods. In terms of NIQE, both $\text{DehazeTab}_{CNN}$ and $\text{DehazeTab}_{Hybrid}$ obtain 3.08, matching the best score achieved by RSVT, while $\text{DehazeTab}_{H}$ and $\text{DehazeTab}_{ViT}$ achieve comparable scores of 3.10 and 3.11, respectively. For BRISQUE, $\text{DehazeTab}_{H}$ obtains the lowest score among the proposed variants (14.56), followed closely by $\text{DehazeTab}_{Hybrid}$ (14.59). Although UCL-Dehaze achieves the best overall BRISQUE score (11.24), the proposed methods outperform several competing approaches, including DCP, BCCR, and USID-Net, on this metric, suggesting that DehazeTab can produce perceptually competitive restorations on natural hazy scenes.

\subsection{Qualitative Comparisons}
\label{subsec:qualitative_results}

Figures~\ref{fig:results_sots} and~\ref{fig:results_hsts} present qualitative comparisons on the SOTS-Outdoor and HSTS datasets, respectively. Overall, DehazeTab effectively removes haze while maintaining natural color reproduction and preserving scene structures, producing results visually close to the ground truths. In particular, the proposed method improves visibility and contrast in distant regions, including buildings, vegetation, and road scenes. Compared with DCP, which often introduces excessive contrast and color distortion, DehazeTab provides more balanced restoration. Although several learning-based methods achieve strong haze suppression, some exhibit over-enhancement, unnatural darkening, or noticeable color shifts. In contrast, DehazeTab generally preserves more consistent illumination and natural colors, although slight residual haze remains in certain distant and low-contrast regions.

Figure~\ref{fig:results_natural} further evaluates generalization to the set of natural hazy scenes. DehazeTab improves scene visibility while largely preserving the original color characteristics and avoiding the strong saturation or contrast enhancement exhibited by some competing approaches. However, residual haze remains noticeable in challenging regions, particularly under dense or spatially complex atmospheric conditions. Overall, these comparisons suggest that the proposed in-context regression framework can achieve visually plausible dehazing across synthetic and natural scenes, while highlighting the remaining challenges in handling severe haze and recovering fine details.

\begin{table}[t]
\centering
\caption{Analysis of different patch sampling methods.}
\resizebox{\columnwidth}{!}{%
\begin{tabular}{lcccccc}
\toprule
\multirow{2}{*}{\text{Method}} & \multirow{2}{*}{\text{Sampling}} & \multicolumn{2}{c}{\text{SOTS-Outdoor}} & \multicolumn{2}{c}{\text{HSTS}} \\
\cmidrule(r){3-4} \cmidrule(r){5-6}
& & \text{PSNR} & \text{SSIM} & \text{PSNR} & \text{SSIM} \\
\midrule
DehazeTab$_{H}$ & Informative & 24.71 & 0.940 & 23.45 & 0.918 \\
DehazeTab$_{CNN}$ & Informative & 23.74 & 0.898 & 23.27 & 0.893 \\
DehazeTab$_{ViT}$ & Informative & 23.06 & 0.884 & 21.18 & 0.858 \\
DehazeTab$_{Hybrid}$ & Informative & 24.87 & 0.942 & 23.36 & 0.916 \\
\midrule
DehazeTab$_{H}$ & Clustering & 24.85 & 0.940 & 23.29 & 0.908 \\
DehazeTab$_{CNN}$ & Clustering & 24.01 & 0.908 & 22.91 & 0.880 \\
DehazeTab$_{ViT}$ & Clustering & 22.92 & 0.882 & 21.06 & 0.849 \\
DehazeTab$_{Hybrid}$ & Clustering & 25.13 & 0.943 & 23.37 & 0.910 \\
\bottomrule
\end{tabular}}
\label{tab:ablation_top_patches}
\end{table}

\subsection{Ablation Studies}
\label{subsec:additional_analysis}

The default configuration (baseline) uses informativeness-based patch selection, retrieves five reference images ($R=5$), and samples 100 patches from each reference ($P=100$). Ablation studies are conducted to assess the impact of these choices by varying each configuration independently while keeping the underlying regression framework unchanged.

\begin{table}[t]
\centering
\caption{Analysis of different numbers of reference images.}
\resizebox{\columnwidth}{!}{%
\begin{tabular}{lcccccc}
\toprule
\multirow{2}{*}{\text{Method}} & \multirow{2}{*}{\text{\#References}} & \multicolumn{2}{c}{\text{SOTS-Outdoor}} & \multicolumn{2}{c}{\text{HSTS}} \\
\cmidrule(r){3-4} \cmidrule(r){5-6}
& & \text{PSNR} & \text{SSIM} & \text{PSNR} & \text{SSIM} \\
\midrule
DehazeTab$_{H}$ & 5 & 24.71 & 0.940 & 23.45 & 0.918 \\
DehazeTab$_{CNN}$ & 5 & 23.74 & 0.898 & 23.27 & 0.893 \\
DehazeTab$_{ViT}$ & 5 & 23.06 & 0.884 & 21.18 & 0.858 \\
DehazeTab$_{Hybrid}$ & 5 & 24.87 & 0.942 & 23.36 & 0.916 \\
\midrule
DehazeTab$_{H}$ & 7 & 24.79 & 0.942 & 23.38 & 0.919 \\
DehazeTab$_{CNN}$ & 7 & 24.06 & 0.903 & 23.49 & 0.904 \\
DehazeTab$_{ViT}$ & 7 & 23.13 & 0.885 & 20.94 & 0.866 \\
DehazeTab$_{Hybrid}$ & 7 & 24.99 & 0.944 & 23.51 & 0.924 \\
\midrule
DehazeTab$_{H}$ & 10 & 25.10 & 0.946 & 24.84 & 0.934 \\
DehazeTab$_{CNN}$ & 10 & 24.27 & 0.910 & 24.17 & 0.915 \\
DehazeTab$_{ViT}$ & 10 & 23.28 & 0.890 & 21.58 & 0.883 \\
DehazeTab$_{Hybrid}$ & 10 & 25.08 & 0.946 & 24.76 & 0.934 \\
\bottomrule
\end{tabular}}
\label{tab:ablation_num_references}
\end{table}

\subsubsection{Effect of Patch Selection Strategy}

Two strategies are compared: informativeness-based selection and feature-space clustering. The informativeness score ranks candidates by grayscale contrast, intensity range, gradient magnitude, entropy, dark-channel variation, and saturation variation, and discourages nearby duplicate patches through a minimum 12-pixel center-distance constraint; whereas the clustering-based approach employs K-Means to partition patch descriptors into clusters and select examples closest to the cluster centers.

Table~\ref{tab:ablation_top_patches} shows that neither strategy consistently dominates. On SOTS-Outdoor, clustering improves the hybrid variant from 24.87 to 25.13 dB and the CNN variant from 23.74 to 24.01 dB, while the ViT variant decreases slightly. On HSTS, informativeness-based selection generally gives stronger SSIM and often stronger PSNR; the hybrid variants achieve nearly identical PSNR (23.36 versus 23.37 dB) but different SSIM (0.916 versus 0.910). These results suggest that patch representativeness and local informativeness capture complementary aspects of the reference distribution, with their usefulness depending on both the representation and dataset.

\subsubsection{Effect of the Number of Reference Images}

Different numbers of retrieved references are tested, $R=\{5,7,10\}$, with 100 patches per image ($P=100$). This controls contextual diversity and table size while keeping the remaining pipeline unchanged. As shown in Table~\ref{tab:ablation_num_references}, increasing the reference count generally benefits restoration, particularly on HSTS. For example, DehazeTab$_H$ improves from 23.45/0.918 with five references to 24.84/0.934 with ten, while DehazeTab$_{Hybrid}$ improves from 23.36/0.916 to 24.76/0.934. On SOTS-Outdoor, the corresponding gains are smaller. Handcrafted features increase from 24.71 to 25.10 dB and hybrid features from 24.87 to 25.08 dB. 
Accordingly, five references ($R=5$) are retained as the default to maintain a comparatively compact inference context, while ten references ($R=10$) provide an effectiveness-oriented alternative.

\subsubsection{Effect of the Number of Patches per Reference}

To assess contextual density, multiple numbers of selected patches per reference are examined, $P=\{50, 100, 150\}$, using five references ($R=5$). Table~\ref{tab:ablation_num_patches} indicates that larger patch sets improve SOTS-Outdoor PSNR for the handcrafted, CNN, and hybrid representations, with the hybrid variant rising from 24.36 dB at 50 patches to 25.23 dB at 150. However, the HSTS results do not show the same trend, handcrafted and hybrid variants reach their highest PSNR at 100 patches, whereas CNN reaches its highest PSNR at 50 patches. Thus, increasing contextual density can add useful examples but may also introduce redundancy or less relevant patches.
The default setting of 100 patches thus provides a reasonable trade-off across the two evaluated datasets.

\begin{table}[t]
\centering
\caption{Analysis of different numbers of patches per image.}
\resizebox{\columnwidth}{!}{%
\begin{tabular}{lcccccc}
\toprule
\multirow{2}{*}{\text{Method}} & \multirow{2}{*}{\text{\#Patches}} & \multicolumn{2}{c}{\text{SOTS-Outdoor}} & \multicolumn{2}{c}{\text{HSTS}} \\
\cmidrule(r){3-4} \cmidrule(r){5-6}
& & \text{PSNR} & \text{SSIM} & \text{PSNR} & \text{SSIM} \\
\midrule
DehazeTab$_{H}$ & 50 & 24.33 & 0.934 & 22.82 & 0.900 \\
DehazeTab$_{CNN}$ & 50 & 23.49 & 0.890 & 23.36 & 0.883 \\
DehazeTab$_{ViT}$ & 50 & 22.96 & 0.885 & 20.89 & 0.851 \\
DehazeTab$_{Hybrid}$ & 50 & 24.36 & 0.934 & 23.10 & 0.902 \\
\midrule
DehazeTab$_{H}$ & 100 & 24.71 & 0.940 & 23.45 & 0.918 \\
DehazeTab$_{CNN}$ & 100 & 23.74 & 0.898 & 23.27 & 0.893 \\
DehazeTab$_{ViT}$ & 100 & 23.06 & 0.884 & 21.18 & 0.858 \\
DehazeTab$_{Hybrid}$ & 100 & 24.87 & 0.942 & 23.36 & 0.916 \\
\midrule
DehazeTab$_{H}$ & 150 & 25.03 & 0.942 & 23.28 & 0.913 \\
DehazeTab$_{CNN}$ & 150 & 24.12 & 0.908 & 23.35 & 0.896 \\
DehazeTab$_{ViT}$ & 150 & 23.12 & 0.885 & 21.05 & 0.860 \\
DehazeTab$_{Hybrid}$ & 150 & 25.23 & 0.944 & 23.29 & 0.914 \\
\bottomrule
\end{tabular}}
\label{tab:ablation_num_patches}
\end{table}

\subsubsection{Inference Time}

Table~\ref{tab:inference_time} compares the per-image processing time of DehazeTab$_{ \{Variant,FeatureDim\} }$ under different numbers of selected patches ($P$) and reference images ($R$). The reported time measures the entire per-image processing pipeline including query-image feature extraction, reference retrieval, TabPFN initialization and prediction, transmission reconstruction, atmospheric-light estimation, and haze-free image recovery. Initial feature-encoder loading and offline table data preparation are excluded.

At the smallest context size ($P=50$, $R=5$, 250 rows), DehazeTab$_H$ requires 2.718~s, compared with 1.930~s for DehazeTab$_{CNN}$ and 2.124~s for DehazeTab$_{ViT}$. Although handcrafted features have lower dimensionality, their extraction involves computing multiple texture statistics for individual patches. In contrast, the CNN and ViT variants obtain patch representations from whole-image encoder feature maps. The resulting difference in query-feature extraction overhead can outweigh the benefit of a smaller tabular input when the context is limited. The hybrid variant, which combines handcrafted descriptors with CNN embeddings, has the highest time at this setting (3.155~s).

\begin{figure}
    \hspace{0.05cm}
    \includegraphics[width=0.45\textwidth]{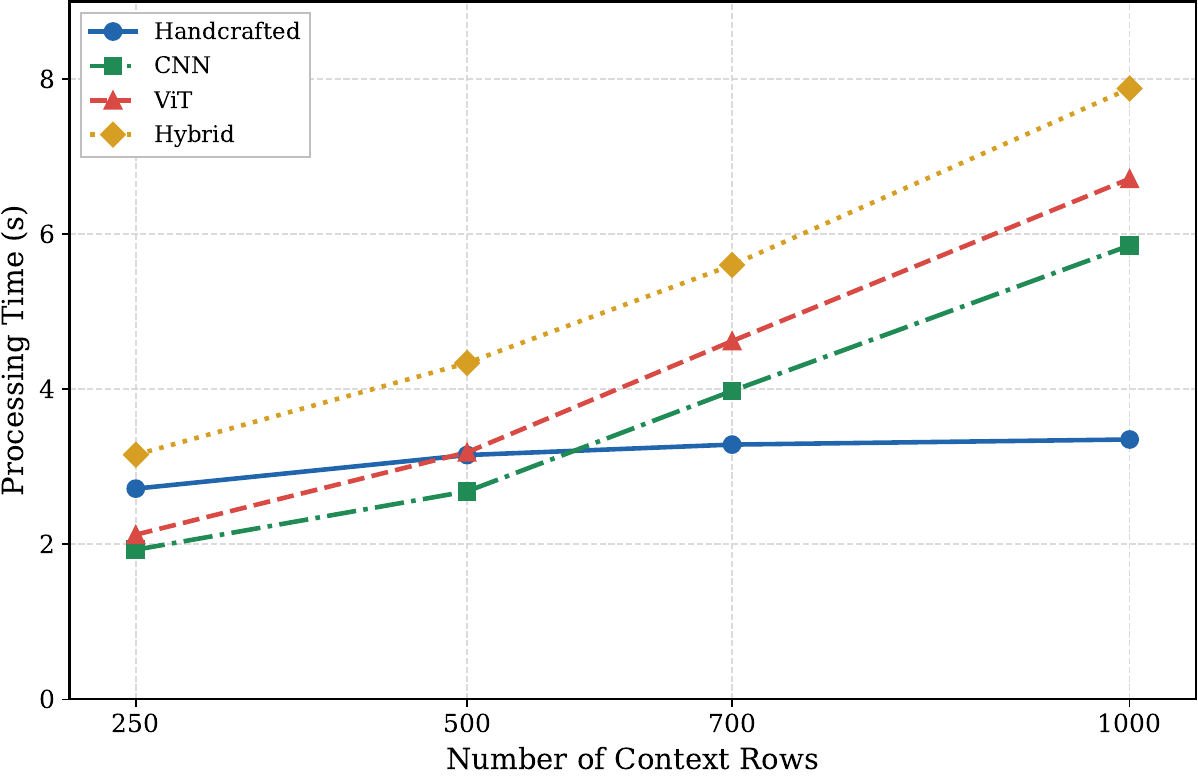}
    \caption{Inference time comparison with different configurations.}
    \label{fig:inference_time}
\end{figure}

As the context grows, the relative cost shifts toward TabPFN processing. With $P=100$, increasing $R$ from 5 to 10 (500 to 1000 context rows) changes the processing time from 3.151 to 3.352~s for DehazeTab$_H$, but from 2.683 to 5.854~s for DehazeTab$_{CNN}$, from 3.186 to 6.714~s for DehazeTab$_{ViT}$, and from 4.338 to 7.879~s for DehazeTab$_{Hybrid}$. This trend is consistent with the greater computational demands of higher-dimensional inputs during TabPFN preprocessing and in-context regression, although the end-to-end measurements do not isolate the contribution of each stage. Increasing $P$ from 100 to 150 at $R=5$ similarly raises the context size from 500 to 750 rows and increases runtime for all variants. Overall, handcrafted features exhibit the most favorable runtime scaling as contextual data increase, while the hybrid representation incurs the largest processing cost. These results highlight a trade-off between feature richness, restoration quality, and end-to-end computational efficiency. Figure \ref{fig:inference_time} visually illustrates the quantitative results reported in Table \ref{tab:inference_time}.

\begin{table}[t]
\centering
\caption{Analysis of inference time with different configurations (in seconds).}
\begin{tabular}{lccccc}
\toprule
\text{\#Patches} ($P$) & 50 & 100 & 100 & 100 & 150 \\
\text{\#References} ($R$) & 5 & 5 & 7 & 10 & 5 \\
\text{\#Context Rows} & 250 & 500 & 700 & 1000 & 750 \\
\midrule
DehazeTab$_{H,44}$ & 2.718 & 3.151 & 3.286 & 3.352 & 3.248 \\
DehazeTab$_{CNN,128}$ & 1.930 & 2.683 & 3.978 & 5.854 & 4.238 \\
DehazeTab$_{ViT,192}$ & 2.124 & 3.186 & 4.621 & 6.714 & 4.896 \\
DehazeTab$_{Hybrid,172}$ & 3.155 & 4.338 & 5.601 & 7.879 & 5.677 \\
\bottomrule
\end{tabular}
\label{tab:inference_time}
\end{table}

\subsection{Discussion and Limitations}
\label{subsec:discussion_limitations}

In summary, the results demonstrate that image-derived structured features can support competitive transmission estimation through a pretrained TFM without dehazing-specific weight updates. The unexpectedly strong performance of handcrafted descriptors highlights the value of haze-aware inductive biases, whereas the mixed benefits of CNN, ViT, and hybrid representations caution against assuming that increasingly complex features automatically improve low-level restoration. The ablations further reveal that contextual reference selection is integral to the approach: the number, diversity, and relevance of examples can materially change reconstruction quality.

On the other hand, several limitations remain. First, transmission targets depend on an estimated atmospheric light and on the atmospheric scattering model, so inaccuracies in these components can propagate into the reconstruction. Second, patch-level scalar predictions may lose fine spatial variation, particularly in dense or spatially nonuniform haze. Finally, in-context inference requires storing and retrieving labeled reference patches, and its computational cost depends on context size and feature extraction.

\section{Conclusions}
\label{sec:conclusions}

This paper introduces DehazeTab, a framework that investigates single-image dehazing from the perspective of in-context tabular inference. By reformulating atmospheric transmission estimation as patch-level structured regression, DehazeTab enables a pretrained TFM to predict transmission values from contextual examples without task-specific model training. 
Experimental results demonstrate the feasibility of this formulation and provide several insights. Handcrafted haze-related features consistently outperform pretrained visual embeddings, highlighting the importance of physically meaningful local information for transmission estimation. CNN-based embeddings outperform ViT-based representations, suggesting that local spatial cues may be more relevant than global context in this setting. Hybrid features offer modest improvements, while ablation studies further reveal the influence of contextual reference and patch sampling on restoration quality.

Although DehazeTab remains limited by its dependence on contextual examples and the challenges of estimating transmission under complex scenes, the findings demonstrate the potential of TFMs beyond conventional tabular prediction. Rather than pursuing state-of-the-art performance, this study provides a preliminary investigation into reformulating low-level vision problems as structured in-context regression, opening opportunities for research on more effective representations, inference strategies, and broader applications.

\ifCLASSOPTIONcaptionsoff
  \newpage
\fi

\balance
\bibliographystyle{IEEEtran}
\bibliography{ref.bib}

\end{document}